\documentclass[11pt]{article}

\usepackage{indentfirst}
\usepackage{float}
\usepackage[margin=1in]{geometry}
\usepackage[T1]{fontenc}
\usepackage[utf8]{inputenc}
\usepackage{lmodern}
\usepackage{amsmath,amssymb}
\usepackage{booktabs,array}
\usepackage{graphicx}
\usepackage{xcolor}
\usepackage{tabularx}
\usepackage{xltabular}
\usepackage{array}    
\usepackage{caption}
\usepackage{enumitem}
\usepackage[numbers,sort&compress]{natbib}
\usepackage{hyperref}
\hypersetup{colorlinks=true,linkcolor=black,citecolor=blue!50!black,urlcolor=blue!50!black}
\title{
\textbf{
How Ground Truth Shapes Retrieval Scores:
Evidence-Aware Evaluation of a Retrieval-Augmented
Generation System for Breast Lesions
}
}

\author{%
Miguel Mochizuki Silva$^{1}$ \qquad
Samuel Ismael$^{1}$ \qquad
Mateus C.\ Dias$^{1}$\\[0.5em]
Thais Gaud\^encio$^{1}$ \qquad
Yuri Malheiros$^{1}$ \qquad
Bruno Barufaldi$^{2,*}$\\[0.8em]
{\small $^{1}$Universidade Federal da Para\'iba (UFPB), Jo\~ao Pessoa, PB, Brazil}\\[0.3em]
{\small $^{2}$Department of Radiology, Perelman School of Medicine,}\\
{\small University of Pennsylvania, Philadelphia, PA, USA}\\[0.3em]
{\small $^{*}$Corresponding author:
\href{mailto:bruno.barufaldi@pennmedicine.upenn.edu}{bruno.barufaldi@pennmedicine.upenn.edu}}%
}

\begin{document}
\maketitle 


\begin{abstract} 

\textbf{Purpose:} Retrieval scores for clinical retrieval-augmented generation (RAG) are usually reported against annotated chunk per queries, which mixes retriever quality with annotation coverage. To describe the corpus, RAG 
(RAG) pipeline, and evidence-aware retrieval evaluation protocol developed 
for a question-answering system over the breast lesion management literature and to report retrieval performance under the selected configuration. 

\textbf{Methods:} An 18-document corpus focused on breast lesions was used for
retrieval benchmarking, ground-truth construction, and generation evaluation.
One simple and direct evaluation query was created for each document, resulting
in an 18-query development benchmark composed of Yes/No and WH-questions.
Retrieval combines dense embeddings, BM25, and a cross-encoder reranker. Chunk size 
(128--1024 tokens), chunk overlap (0--20\%) varied one factor at time, and four retriever
configurations were compared in an ablation. Ground-truth 
relevance is defined at the level of evidence spans, which are mapped to 
every chunk containing them, so that chunks expressing the same underlying 
fact under overlapping chunking are scored as equivalent rather than as 
competing candidates. Retrieval quality is reported with R-Precision, Recall@K, 
MRR, and nDCG@K on an 18-query ground-truth development set. Generation 
quality was assessed in a separate experiment on an 512-token index, using DeepEval claim-level metrics with 
\texttt{deepseek-v4-pro} as the LLM judge, together with classical BLEU/ROUGE 
metrics across context depths K = 1–5.

\textbf{Results:} The selected hybrid retriever (384-token chunks, 20\% overlap) retrieved accepted evidence in the top three results for 17 of 18 queries. Recall@3, R-Precision, MRR@3, and nDCG@3 each reached 0.94. Ground-truth curation increased the number of evidence spans from 25 to 37 and accepted chunk targets from 36 to 51 without changing retrieval rankings.

\textbf{Conclusion:} Cross-encoder reranking produced the highest retrieval scores. Ground-truth curation expanded the set of recognized evidence without changing retrieval rankings, underscoring the dependence of evaluation scores on relevance annotations. The small development set and retrieval-informed curation warrant independent validation, including radiologist assessment of evidence sufficiency.

\end{abstract}
\bigskip

\section{Introduction}

Management of benign and high-risk breast lesions requires integrating imaging findings, biopsy pathology, radiologic--pathologic concordance, and patient risk. Supporting evidence is distributed across society guidelines, appropriateness criteria, and lesion-specific studies, which may differ in their recommendations. For lesions of uncertain malignant potential (B3 lesions), international consensus supports vacuum-assisted excision as an alternative to open surgery for selected categories \citep{rageth2019b3}, whereas U.S. guidance permits imaging surveillance for selected radiologic--pathologic concordant lesions \citep{asbrs2024highrisk}. Retrieving the correct recommendation therefore requires preserving the clinical conditions under which it applies.

Large language models (LLMs) encode substantial clinical knowledge \citep{singhal2023medpalm} but may generate unsupported statements \citep{ji2023hallucination}. Retrieval-augmented generation (RAG) conditions answers on retrieved passages and makes supporting sources available for inspection \citep{lewis2020rag}. In medical question answering, retrieval has improved performance in some settings, although outcomes depend on the corpus and retrieval strategy \citep{xiong2024medrag}. Missing evidence can limit answer correctness, while supplying additional passages does not necessarily resolve the problem because language models may fail to use relevant information in long contexts \citep{liu2024lost}.

Medical RAG systems have been evaluated using guidelines and clinical resources, including clinical question answering and simulated oncology cases \citep{zakka2023almanac,ferber2025oncologyagent}. Radiology-focused studies have also reported improvements that vary across models and sources \citep{tayebiarasteh2025radiorag,weinert2025radiographics}. However, evaluation protocols remain heterogeneous \citep{amugongo2025ragreview}, and expert assessment has identified retrieval and evidence selection as important failure points \citep{kim2025rethinkingrag}. These findings motivate evaluating retrieved evidence separately from generated answers.

Retrieval scores depend on which passages are annotated as relevant. When a benchmark recognizes only one chunk per question, another chunk containing the same supporting evidence may be scored as nonrelevant. Overlapping segmentation makes such alternatives more likely, while additional relevant passages may remain unannotated. In information retrieval, incomplete relevance judgments have long been recognized as a source of evaluation bias \citep{zobel1998reliable,buettcher2007biased,buckley2004incomplete}. The effects on individual metrics depend on their definitions and the accepted evidence targets; overlap alone does not establish a fixed penalty.

Question-answering benchmarks have represented supporting evidence through answer spans, sentence-level facts, and source provenance \citep{rajpurkar2016squad,kwiatkowski2019naturalquestions,yang2018hotpotqa,petroni2021kilt}. In RAG evaluation, conventional ranking metrics may not fully capture whether retrieved passages supply the information needed to answer a question \citep{yu2024ragsurvey}. Chunk-based evaluation is also sensitive to how evidence is partitioned and repeated across overlapping segments \citep{smith2024chunking}. Evidence-level relevance annotations can therefore reduce dependence on a single chunk identifier, provided that accepted alternatives are explicitly defined and audited.

We present RAG-MAMA, a retrieval-augmented generation system evaluated on a curated 18-document breast-lesion corpus and an 18-query development benchmark. We investigate whether incomplete ground-truth annotations can underestimate retrieval performance. Canonical evidence spans are defined independently of chunk identifiers and mapped to each segmentation configuration. We distinguish alternative supporting chunks from complementary evidence requirements, audit previously unrecognized evidence while keeping retrieval rankings fixed, and separately evaluate retrieval configurations and generation quality. The system focuses on excision-versus-surveillance decisions, with fixed deferral for excluded categories. This study establishes a small-scale methodological baseline rather than demonstrating clinical effectiveness or validity on more complex, multi-evidence questions.

\section{Methods}

\begin{figure}[H]
    \centering
    \includegraphics[width=\textwidth]{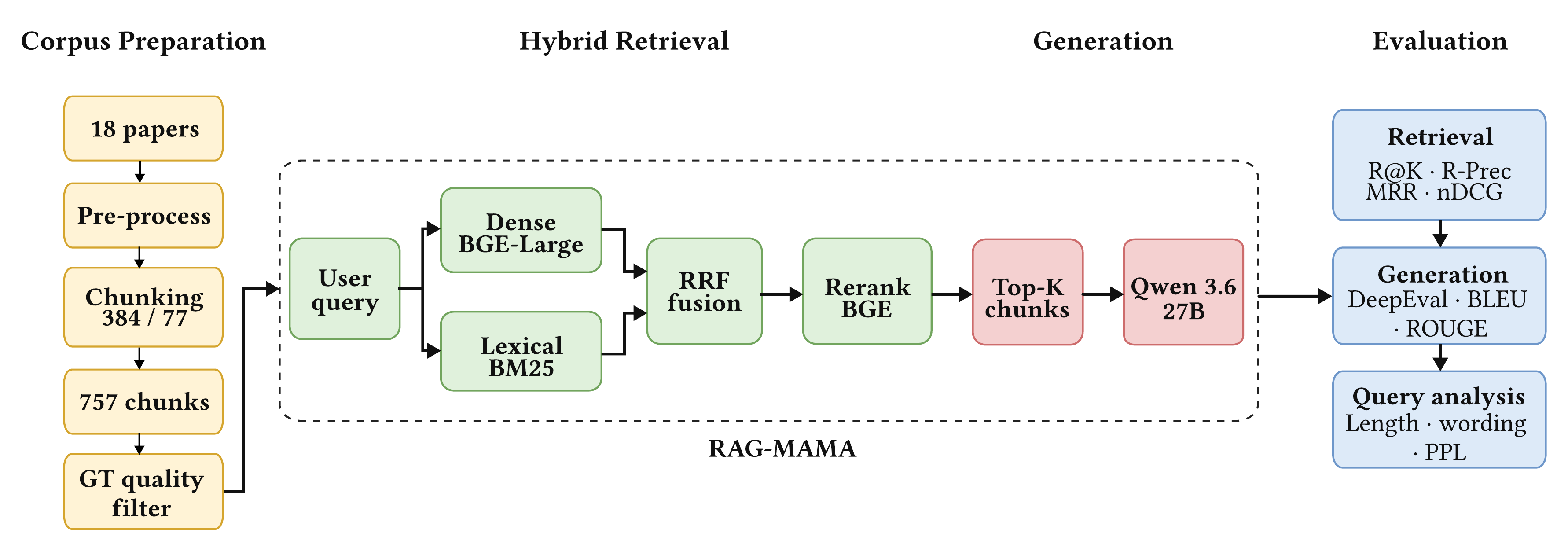}
    \caption{Overview of the RAG-MAMA architecture, illustrating corpus preparation, hybrid retrieval, response generation, and evaluation.}
    \label{fig:rag-architecture}
\end{figure}

Figure~\ref{fig:rag-architecture} shows the RAG-MAMA pipeline. The corpus comprised 18 papers, which were preprocessed and split into 757 chunks (size 384, overlap 77). Each query was retrieved in parallel by a dense retriever (BGE-Large) and a lexical retriever (BM25). The two ranked lists were combined with Reciprocal Rank Fusion and reranked with a BGE cross-encoder. The top-K chunks were passed to Qwen 3.6 (27B) to generate a grounded answer. Retrieval was evaluated with ranking metrics and generation with DeepEval.

\subsection{Corpus Compilation, Pre-processing, and Quality Assurance}
\label{sec:corpus}

The corpus was scoped a priori to a fixed reference table of 15 breast pathology categories defined by the American Society of Breast Surgeons (ASBrS) high-risk lesion taxonomy \citep{asbrs2024highrisk}.

Thirty candidate documents were assembled by purposive expert selection, comprising society guidelines, American College of Radiology (ACR) appropriateness criteria, World Health Organization (WHO) classification, and lesion-specific primary literature. Documents were eligible if they provided management recommendations, upgrade-risk estimates, radiologic–pathologic concordance criteria, or nomenclature definitions for at least one retrieval-eligible category. Each was evaluated by category-specific term frequency and adjudicated by full-text review. Malignant categories and fibromatosis/desmoid tumor are not answered by retrieval: a scope gate applied to the normalized pathology label returns a fixed deferral before retrieval runs. Fibromatosis is excluded because current sarcoma guidance favors active surveillance over the excision and desmoid text remains in the indexed corpus.

\begingroup
\small
\setlength{\tabcolsep}{4pt}
\renewcommand{\arraystretch}{1.35}
\setlength{\extrarowheight}{1.5pt}

\begin{xltabular}{\textwidth}{
>{\raggedright\arraybackslash}X
>{\raggedright\arraybackslash}p{2.4cm}
>{\raggedright\arraybackslash}p{2.3cm}
>{\raggedright\arraybackslash}p{2.7cm}}

\caption{Classification and typical management of benign, high-risk, and malignant breast pathology entities. Malignant lesions are routed to a fixed response by the scope gate.}
\label{tab:pathology_classification}\\
\toprule
\textbf{Pathology entities} & \textbf{Classification} & \textbf{ASBrS category} & \textbf{Typical management}\\
\midrule
\endfirsthead

\multicolumn{4}{l}{\textit{Table \thetable\ (continued)}}\\
\toprule
\textbf{Pathology entities} & \textbf{Classification} & \textbf{ASBrS category} & \textbf{Typical management}\\
\midrule
\endhead

\bottomrule
\endfoot

Atypical ductal hyperplasia (ADH) & High-risk lesion & ADH & Surgical excision\\
Flat epithelial atypia (FEA) & High-risk lesion & Flat epithelial atypia & Selective excision or observation\\
Lobular carcinoma in situ (LCIS); atypical lobular hyperplasia (ALH) & High-risk lesion & Lobular neoplasia & Observation or selective excision\\
Pleomorphic LCIS; florid LCIS & High-risk lesion & Non-classic LCIS & Surgical excision\\
Papilloma without atypia; intraductal papilloma; peripheral duct papilloma & Benign lesion & Papillary lesions & Observation or selective excision\\
Papilloma with atypia & High-risk lesion & Papillary lesions & Surgical excision\\
Radial scar; complex sclerosing lesion & High-risk lesion & Radial scar / complex sclerosing lesion & Observation or excision\\
Mucocele-like lesion & High-risk lesion & Mucocele-like lesion & Observation or excision\\
Pseudoangiomatous stromal hyperplasia (PASH) & Benign lesion & PASH & Observation in most patients\\
Fibromatosis; desmoid tumor & Benign lesion & Fibromatosis & Deferred: sarcoma multidisciplinary team\\
Sclerosing adenosis; adenosis; apocrine metaplasia; usual ductal hyperplasia; fibroadenomatoid hyperplasia & Benign proliferative lesion & --- & Routine clinical management\\
Fibrocystic change; fibroadenoma; benign cyst; fat necrosis; duct ectasia; hamartoma; and related benign diagnoses & Benign lesion & --- & Routine clinical management\\
Ductal carcinoma in situ (DCIS), all grades and histologic subtypes; intraductal (adeno)carcinoma, non-infiltrating & Malignant (Stage 0) & --- & Breast cancer treatment guidelines\\
Invasive (infiltrating) ductal carcinoma; invasive (infiltrating) lobular carcinoma; mixed ductal and lobular carcinoma; invasive mammary carcinoma; tubular carcinoma; metaplastic carcinoma & Malignant & --- & Breast cancer treatment guidelines\\
Metastatic carcinoma to axillary or intramammary lymph node; lymphoma involving the breast; metastatic melanoma to the breast & Malignant & --- & Disease-specific oncologic management\\

\end{xltabular}
\endgroup

Eighteen documents were retained and indexed, stratified into four tiers used as retrieval priors (Table \ref{tab:guidelines}): core management guidelines (n = 4), lesion-specific and concordance sources (\textit{n} = 9), imaging appropriateness criteria (\textit{n} = 3), and benign breast disease risk literature (\textit{n} = 2). Ten documents were excluded: treatment guidelines for established malignancy (\textit{n} = 3), since the reference table defers these categories without requiring their text and their volume (NCCN with over 820k chars) would dominate retrieval; disease entities outside the table (\textit{n} = 3); documents without lesion-specific management content (\textit{n} = 2); a superseded WHO edition (\textit{n} = 1); and the EUSOMA B3 guideline (\textit{n} = 1), which specifies vacuum-assisted excision as the default therapeutic step whereas the retained guidelines frame management as excision versus surveillance. The retained corpus is therefore internally consistent on the excision-versus-surveillance axis and represents North American practice.


\begingroup
\small
\setlength{\tabcolsep}{3.5pt}
\renewcommand{\arraystretch}{1.35}
\setlength{\extrarowheight}{1.5pt}

\begin{xltabular}{\textwidth}{
>{\raggedright\arraybackslash}p{2.0cm}
>{\raggedright\arraybackslash}X
>{\centering\arraybackslash}p{1.2cm}
>{\raggedright\arraybackslash}p{2.7cm}}

\caption{Guidelines and scientific literature supporting the management of breast lesions.}
\label{tab:guidelines}\\
\toprule
\textbf{Authors} &
\textbf{Source} &
\textbf{Year} &
\textbf{Document type}\\
\midrule
\endfirsthead

\multicolumn{4}{l}{\textit{Table \thetable\ (continued)}}\\
\toprule
\textbf{Authors} &
\textbf{Source} &
\textbf{Year} &
\textbf{Document type}\\
\midrule
\endhead

\bottomrule
\endfoot

ASBrS &
Resource Guide: Surgical Management of Benign or High-Risk Lesions &
2024 &
Archived society resource guide\\

Nakhlis et al. &
American Society of Breast Surgeons, Society of Breast Imaging, and College of American Pathology 2026 Guidelines for the Management of Proliferative Lesions With Atypia and Lobular Carcinoma In Situ. \textit{JAMA Surg} &
2026 &
Society guideline / Special Communication\\

Zamora et al. &
Management of High-Risk Breast Lesions: A Comprehensive Update. \textit{J Breast Imaging} &
2026 &
Scientific review\\

Eisen et al. &
High-Risk Benign Breast Lesions: An Ontario Health (Cancer Care Ontario) Recommendations Report. \textit{Curr Oncol} &
2026 &
Recommendations report / review\\

Rosenberger et al. &
American Society of Breast Surgeons and Society of Breast Imaging 2025 Guidelines for the Management of Benign Breast Fibroepithelial Lesions. \textit{JAMA Surg} &
2025 &
Society guideline / Special Communication\\

Mitchell et al. &
American Society of Breast Surgeons, Society of Breast Imaging, and College of American Pathology 2025 Guidelines for the Management of Infectious and Inflammatory Lesions of the Breast. \textit{JAMA Surg} &
2026 &
Society guideline / Special Communication\\

Guirguis et al. &
Mimickers of Breast Malignancy: Imaging Findings, Pathologic Concordance and Clinical Management. \textit{Insights Imaging} &
2021 &
Educational review\\

Estes et al. &
Pseudoangiomatous Stromal Hyperplasia: Overview and Clinical Management. \textit{Ann Breast Surg} &
2020 &
Review article\\

Yoon et al. &
Optimal Treatment of Pseudoangiomatous Stromal Hyperplasia of the Breast. \textit{Asian J Surg} &
2020 &
Original cohort study\\

Quinn et al. &
World Health Organization Classification of Tumours of the Breast 6th Edition 2026. \textit{Histopathology} &
2026 &
WHO classification review\\

ASBrS &
Concordance Assessment Following Image-Guided Breast Biopsy &
2010 &
Quality measure\\

ASBrS &
Image-Guided Percutaneous Biopsy of Palpable and Nonpalpable Breast Lesions &
2018 &
Society resource guide\\

ASBrS &
Position Statement on Stereotactic Breast Procedures &
2010 (rev.) &
Position / practice guideline\\

Sanford et al. &
ACR Appropriateness Criteria: Evaluation of Nipple Discharge: 2022 Update. \textit{J Am Coll Radiol} &
2022 &
Appropriateness criteria\\

Klein et al. &
ACR Appropriateness Criteria: Palpable Breast Masses: 2022 Update. \textit{J Am Coll Radiol} &
2023 &
Appropriateness criteria\\

Niell et al. &
ACR Appropriateness Criteria: Evaluation of the Symptomatic Male Breast. \textit{J Am Coll Radiol} &
2018 &
Appropriateness criteria\\

Hartmann et al. &
Benign Breast Disease and the Risk of Breast Cancer. \textit{N Engl J Med} &
2005 &
Cohort study\\

Dyrstad et al. &
Breast Cancer Risk Associated With Benign Breast Disease: Systematic Review and Meta-analysis. \textit{Breast Cancer Res Treat} &
2015 &
Systematic review and meta-analysis\\

\end{xltabular}
\endgroup

An 18-query retrieval benchmark was constructed, with one question per source document. Questions, reference answers, and supporting evidence were drafted with Codex (recorded session model: \texttt{gpt-5.6-terra}) and reviewed by the authors. Each Yes/No or WH query targeted a focused, source-supported answer.

Under the selected fixed-size chunking configuration of 384 tokens with a 77-token overlap, the 18-document corpus yielded a total of 757 chunks. Text is extracted with PyMuPDF (1.23.0+). Unicode control characters (U+0000--U+009F) are replaced with spaces, soft hyphens (U+00AD) and zero-width spaces (U+200B) are removed, and whitespace is collapsed. Repeated header/footer lines are removed when they recur three or more times. Reference sections are identified by heading pattern and removed when citation density exceeds 25\% of lines.

A seven-feature noise classifier assigned chunks to five categories. All 757 chunks were stored in ChromaDB; 558 were eligible for retrieval after filtering.

\subsection{Retrieval-Augmented Generation Pipeline}
\label{sec:rag}

The system implements a modular RAG architecture. Dense retrieval uses
\texttt{BAAI/bge-large-en-v1.5} embeddings over a ChromaDB vector store,
while lexical retrieval uses BM25 \citep{robertson2009bm25}. The semantic and
lexical rankings are combined using reciprocal-rank fusion (RRF)
\citep{cormack2009rrf}, with an RRF parameter of 50. The selected environment
configures a hybrid fetch multiplier of 2 and a semantic retrieval cap of
30 results. Up to 20 fused candidates are subsequently re-scored using the
\texttt{BAAI/bge-reranker-v2-m3} cross-encoder
\citep{nogueira2019passage,xiao2023cpack}. The final retrieval depth used in
the initial benchmark is \(K=3\). Table \ref{tab:config} shows the full configuration.

\begin{table}[h]
\centering
\small
\setlength{\tabcolsep}{5pt}
\renewcommand{\arraystretch}{1.30}
\setlength{\extrarowheight}{1.2pt}

\caption{Selected retrieval and generation configuration.}
\label{tab:config}

\begin{tabularx}{\textwidth}{
@{}>{\raggedright\arraybackslash}p{3.2cm}
>{\raggedright\arraybackslash}X@{}}
\toprule
\textbf{Component} & \textbf{Setting} \\
\midrule

Chunking &
Fixed-size, 384 tokens; 77-token (20\%) overlap; 757 chunks \\

Tokenizer for chunking &
\texttt{bge-large-en-v1.5} (Hugging Face \texttt{AutoTokenizer}) \\

\addlinespace[2pt]

Dense retrieval &
\texttt{BAAI/bge-large-en-v1.5}; ChromaDB; semantic cap 30 \\

Lexical retrieval &
BM25 (\texttt{rank-bm25}); biomedical-aware tokenizer \\

Fusion &
RRF, $k=50$; hybrid fetch multiplier 2 \\

Reranker &
\texttt{BAAI/bge-reranker-v2-m3}; candidate pool 20 \\

Final retrieval depth &
$K=3$ \\

\addlinespace[2pt]

Generator &
\texttt{qwen/qwen3.6-27b}; temperature 0.3; max output 2048 tokens; prompt \texttt{rag-evaluation-v1} \\

Generation context &
Top-$K$ chunks in rank order; each capped at 3{,}500 characters; metadata included \\

Judge &
\texttt{deepseek-v4-pro}; DeepEval 4.0.7; threshold 0.5 \\

\bottomrule
\end{tabularx}
\end{table}

\subsubsection{Chunking}

Token counts used during chunk construction were computed using the tokenizer
associated with the dense embedding model,
\texttt{BAAI/bge-large-en-v1.5}, loaded through the Hugging Face
\texttt{AutoTokenizer} interface.

The fixed-size chunking procedure operates on extracted text units rather than
splitting the normalized document at arbitrary character positions. Text units
are packed until adding another unit would exceed the configured token limit.
Overlap between consecutive chunks is constructed from trailing text units
whenever possible. Chunk-level provenance, including page ranges, contributing
text blocks, and section metadata, is retained during this process.

Chunk size and overlap were varied one factor at a time. Chunk size was varied from 128 to 1024 tokens with approximately 20\% overlap. After selecting a chunk size of 384 tokens, overlap was varied from 0\% to 20\%, corresponding to 0, 38, 58, and 77 tokens. The selected configuration uses 384-token chunks with a 77-token overlap.

\subsubsection{Embedding, reranking, and lexical retrieval}

Dense retrieval represents both documents and queries using
\texttt{BAAI/bge-large-en-v1.5}. Lexical retrieval uses BM25 with a separate
biomedical-aware tokenizer.

The BM25 tokenizer lowercases text, normalizes hyphen variants, removes common
punctuation, and preserves clinically relevant lexical forms such as HER2+,
ER-, BI-RADS, percentages, decimal values, and comparison operators.

The semantic and lexical rankings are combined using RRF with parameter 50.
The hybrid retrieval stage uses a fetch multiplier of 2, with the semantic
retrieval stage configured to return up to 30 results. Up to 20 fused
candidates are then re-scored using the
\texttt{BAAI/bge-reranker-v2-m3} cross-encoder, and the top three chunks are
returned as the final retrieval output.

The retrieval ablation compares four configurations: BM25, dense semantic
retrieval, hybrid RRF, and hybrid RRF followed by cross-encoder reranking.
All variants use the same corpus, 18-query development set, selected
384-token/77-token segmentation, and relevance annotations for the ablation;
only the retrieval component under evaluation is changed.

\subsubsection{Answer generation}

The answer-generation component was performed using
\texttt{qwen/qwen3.6-27b} through an OpenAI-compatible chat-completions
interface (Table \ref{tab:config}). Generation context is constructed from the highest-ranked retrieved chunks in
their final retrieval order. For each selected chunk, the implementation keeps
at most the 3,500 characters of chunk text and then adds
identifying metadata, including chunk ID, document ID, source file, DOI, and
rank. Truncation is therefore character-based rather than token-based and may
occur inside a sentence, table, or section.

The implementation does not enforce a separate global token budget or a
project-level model context-window check for answer generation. Chunks below
the configured generation-context depth are omitted, while each selected chunk
is independently capped at 3,500 characters. If the generation request fails,
the run records a generation failure rather than rebuilding a smaller context.
No generation failures occurred in the preserved K1--K5 full-RAG experiments.

\
\subsection{Retrieval Evaluation}
\label{sec:retrieval-eval}

The development benchmark contains 18 queries, one per source document, with
simple and direct Yes/No or WH formulations. Each query is associated with
canonical evidence containing the information required to support its
reference answer.

Retrieval was evaluated using Recall@3, R-Precision, MRR@3, and nDCG@3 \citep{jarvelin2002ndcg}. Results were macro-averaged across 18 queries.

Canonical evidence is maintained independently from chunk identifiers.
Because different chunk-size and overlap configurations produce different
chunk boundaries and identifiers, an evidence-to-chunk binding is generated
for each evaluated collection. Thus, the chunk-size and overlap sweeps use the
same 18-query development ground truth while mapping its canonical evidence to
the chunk representation produced by each segmentation configuration.

Automatic evidence-to-chunk binding matched the complete canonical evidence span to chunks from the same document after Unicode NFC normalization and whitespace collapsing, preserving case and punctuation. When no individual chunk contained the complete span, the mapper examined consecutive chunks, removed exact suffix--prefix token overlap, and identified the shortest sequence reconstructing the span. Partial containment alone did not qualify a chunk as a standalone evidence target. Additional accepted chunks could be included through manual curation; their clinical sufficiency and preservation of required qualifiers were annotation judgments rather than guarantees of string matching.

Evidence routes distinguish alternative chunks (OR) from complementary chunks required jointly (AND). In the reported 18-query benchmark, however, each query had one required evidence group, and all accepted targets were individual chunks. No complementary multi-chunk routes were annotated. Each alternative was evaluated separately, retaining the maximum score per query and metric before macro-averaging. Consequently, Recall@3 indicates whether any accepted alternative appears within the first three ranks; R-Precision indicates an accepted rank-one result; MRR@3 uses the reciprocal of the best accepted rank up to three; and nDCG@3 uses discounted binary gain. Equivalent chunks were not counted as multiple relevant items within a single target, but the scores can still depend on the accepted alternatives and their positions. Invariance to overlapping chunks is therefore not established.

Following selection of 384-token chunks with 77-token overlap, high-ranking chunks absent from the initial ground truth were reviewed, including those ranked above the original gold evidence. Additional chunks judged to support the reference answers were added to the curated relevance set. The final retrieval ablation used these annotations. Rankings remained fixed during ground-truth enrichment; score changes therefore reflect revised relevance labels, not retriever changes.

An exploratory analysis of query length, wording, question form, and perplexity is reported in Supplementary Section~\ref{sec:supp-query-formulation}.

\subsection{Generation Evaluation}
\label{sec:gen-eval}

Generation evaluation separates retrieval from answer generation. The
preserved full-RAG experiment uses the same 18 benchmark queries and keeps a
fixed Top-5 retrieval result for each query. Generation-context depth is then
varied from \(K=1\) to \(K=5\): the first one, two, three, four, or five
retrieved chunks, respectively, are supplied to the generator. Thus, \(K\) in
this experiment denotes generation-context depth rather than a change in the
underlying retrieval depth.

These archived generation runs are associated with an earlier index of the
same 18-document corpus, using fixed 512-token chunks with a 102-token overlap
and yielding 560 chunks. This differs from the subsequently selected
384-token/77-token configuration used for the retrieval segmentation and
ablation experiments. All five generation-context conditions preserve the
same ordered Top-5 retrieval results and vary only the number of those chunks
passed to answer generation.

For each selected chunk, at most 3500 characters of chunk text are supplied,
so the maximum chunk-text contribution is 3500, 7000, 10500, 14000, and
17500 characters for K1 through K5, respectively, before metadata and prompt
overhead. These values are derived per-condition bounds rather than enforced
global context limits. No tokenizer-based generation-context budget is
applied. All K1--K5 runs use the recorded model ID
\texttt{qwen/qwen3.6-27b}, temperature 0.3, maximum output length of
2048 tokens, and prompt version \texttt{rag-evaluation-v1}. 

Traditional answer evaluation metrics compare each generated answer with its
\texttt{reference\_answer}. BLEU is computed with SacreBLEU 2.6.0 using the
13a tokenizer, lowercasing, Unicode NFC normalization, exponential smoothing,
and effective order; reported BLEU-1 through BLEU-4 values are cumulative
sentence BLEU-N scores averaged across queries. ROUGE is computed with
rouge-score 0.1.2 using its standard tokenizer, stemming enabled, and
F-measure for ROUGE-1, ROUGE-2, and ROUGE-L
\citep{papineni2002bleu,lin2004rouge}.

Model-based evaluation uses \texttt{DeepEval 4.0.7} with the recorded judge model
\texttt{deepseek-v4-pro} through an OpenAI-compatible DeepSeek endpoint,
synchronous execution, and a threshold of 0.5. The evaluated
metrics are Contextual Precision, Contextual Recall, Contextual Relevancy,
Faithfulness, and Answer Relevancy.

\section{Results}
\label{sec:results}

With 384-token chunks and 20\% overlap, hybrid retrieval followed by cross-encoder reranking retrieved accepted evidence among the top three results for 17 of the 18 development queries (Recall@3 = 0.94). Ground-truth curation increased the number of accepted evidence spans from 25 to 37 without changing retrieval rankings. Generation performance varied across context depths.

\subsection{Segmentation}
\label{sec:res-seg}

\begin{figure}[htbp]
    \centering

    \begin{minipage}[t]{0.49\textwidth}
        \centering
        \includegraphics[width=\textwidth]{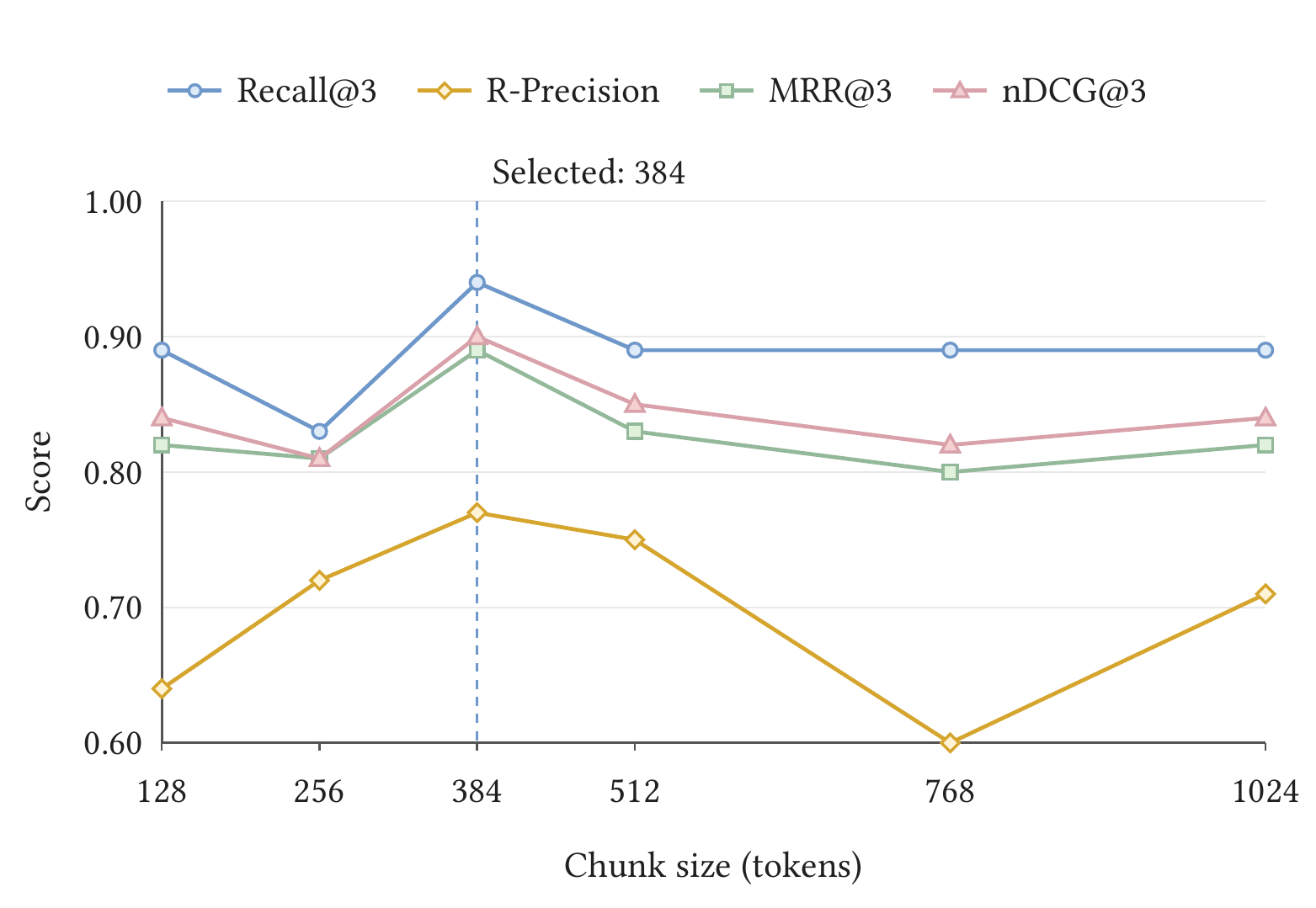}
        \par\smallskip
        {\small (A) Chunk-size sweep}
    \end{minipage}
    \hfill
    \begin{minipage}[t]{0.49\textwidth}
        \centering
        \includegraphics[width=\textwidth]{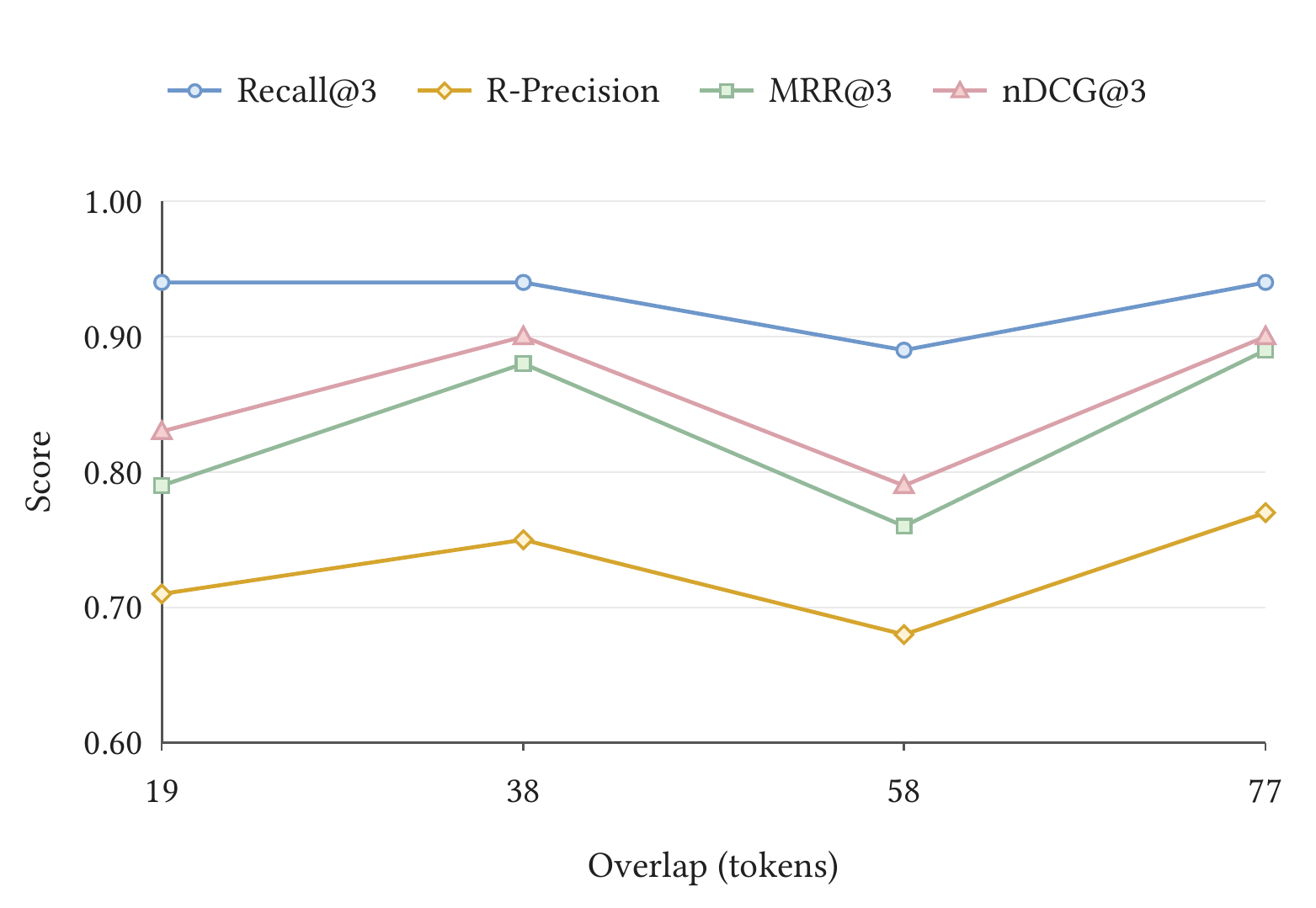}
        \par\smallskip
        {\small (B) Overlap sweep}
    \end{minipage}

    \caption{Retrieval performance under different segmentation configurations on the 18-query development benchmark. (A) Chunk-size sweep with approximately 20\% overlap. (B) Overlap sweep at a fixed chunk size of 384 tokens.}
    \label{fig:segmentation-comparison}
\end{figure}

Figure~\ref{fig:segmentation-comparison}(A) compares chunk sizes from 128 to 1024 tokens at approximately 20\% overlap. The 384-token configuration achieved the highest overall retrieval performance (Recall@3 = 0.94, R-Precision = 0.77, MRR@3 = 0.89, and nDCG@3 = 0.90).

At a fixed chunk size of 384 tokens, Recall@3 remained at 0.94 for 0\%, 10\%, and 20\% overlap (Figure~\ref{fig:segmentation-comparison}(B)). The 20\% setting (77 tokens) achieved the highest R-Precision (0.77), MRR@3 (0.89), and nDCG@3 (0.90) and was selected for subsequent retrieval evaluations.

\subsection{Retrieval Ablation}
\label{sec:res-ablation}

Table~\ref{tab:retrieval-ablation} compares four retrieval configurations against the curated ground truth. BM25 and semantic retrieval achieved Recall@3 values of 0.89 and 0.83, respectively. Hybrid RRF maintained Recall@3 at 0.89 while increasing R-Precision to 0.78. Adding the cross-encoder reranker increased Recall@3, R-Precision, MRR@3, and nDCG@3 to 0.94.

\begin{table}[htbp]
\centering
\small
\setlength{\tabcolsep}{3pt}
\caption{Retrieval ablation on the 18-query development set after ground-truth curation. $\rightarrow 1$ indicates that values closer to 1 are better; bold values indicate the best observed scores.}
\label{tab:retrieval-ablation}

\begin{tabular*}{\textwidth}{@{\extracolsep{\fill}}lcccc@{}}
\toprule
Retriever & Recall@3 $\rightarrow 1$ & R-Precision $\rightarrow 1$ & MRR@3 $\rightarrow 1$ & nDCG@3 $\rightarrow 1$ \\
\midrule
BM25                  & 0.89 & 0.61 & 0.72 & 0.76 \\
Semantic              & 0.83 & 0.56 & 0.69 & 0.73 \\
Hybrid RRF            & 0.89 & 0.78 & 0.81 & 0.83 \\
Hybrid RRF + Reranker & \textbf{0.94} & \textbf{0.94} & \textbf{0.94} & \textbf{0.94} \\
\bottomrule
\end{tabular*}

\end{table}

\subsection{Effect of Ground-Truth Curation}
\label{sec:res-curation}

Ground-truth curation increased the number of evidence spans from 25 to 37 and accepted chunk targets from 36 to 51 without changing retrieval rankings. Against the curated annotations, Hybrid RRF + Reranker retrieved an accepted evidence target within the top three for 17 of 18 queries (Recall@3 = 0.94). R-Precision, MRR@3, and nDCG@3 also reached 0.94.

\subsection{Generation Quality by Context Depth}
\label{sec:res-generation}

Table~\ref{tab:generation-evaluation}(A) presents the BLEU and ROUGE scores across generation-context depths from \(K=1\) to \(K=5\). The highest BLEU-1 through BLEU-4, ROUGE-1, and ROUGE-2 scores occurred at \(K=5\), including BLEU-4 = 0.22, ROUGE-1 = 0.56, and ROUGE-2 = 0.37. ROUGE-L was highest at \(K=1\) (0.51).

Table~\ref{tab:generation-evaluation}(B) presents the DeepEval metrics. Contextual Relevancy decreased from 0.72 at \(K=1\) to 0.47 at \(K=5\), while Contextual Recall reached 0.94 at \(K=2\), \(K=4\), and \(K=5\). Contextual Precision was highest at \(K=5\) (0.91). Faithfulness remained between 0.96 and 1.00, and Answer Relevancy reached 0.94 at \(K=2\), \(K=4\), and \(K=5\). No single context depth consistently maximized all metrics.

\begin{table}[htbp]
\centering
\small
\caption{Generation evaluation across context depths K1--K5. (A) Classical BLEU/ROUGE metrics. (B) DeepEval metrics. $\rightarrow 1$ indicates that values closer to 1 are better; bold values indicate the best observed scores before rounding.}
\label{tab:generation-evaluation}

\vspace{0.4em}
\textbf{(A) Classical generation metrics}

\vspace{0.4em}
\begin{tabular*}{\textwidth}{@{\extracolsep{\fill}}lccccc@{}}
\toprule
Metric & K1 & K2 & K3 & K4 & K5 \\
\midrule
BLEU-1 $\rightarrow 1$  & 0.44 & 0.41 & 0.43 & 0.38 & \textbf{0.44} \\
BLEU-2 $\rightarrow 1$  & 0.33 & 0.31 & 0.32 & 0.29 & \textbf{0.34} \\
BLEU-3 $\rightarrow 1$  & 0.26 & 0.24 & 0.25 & 0.23 & \textbf{0.27} \\
BLEU-4 $\rightarrow 1$  & 0.22 & 0.19 & 0.20 & 0.18 & \textbf{0.22} \\
ROUGE-1 $\rightarrow 1$ & 0.55 & 0.54 & 0.54 & 0.50 & \textbf{0.56} \\
ROUGE-2 $\rightarrow 1$ & 0.36 & 0.35 & 0.34 & 0.30 & \textbf{0.37} \\
ROUGE-L $\rightarrow 1$ & \textbf{0.51} & 0.47 & 0.47 & 0.42 & 0.50 \\
\bottomrule
\end{tabular*}

\vspace{1em}
\textbf{(B) DeepEval metrics}

\vspace{0.4em}
\begin{tabular*}{\textwidth}{@{\extracolsep{\fill}}lccccc@{}}
\toprule
Metric & K1 & K2 & K3 & K4 & K5 \\
\midrule
Contextual Precision $\rightarrow 1$ & 0.89 & 0.89 & 0.88 & 0.87 & \textbf{0.91} \\
Contextual Recall $\rightarrow 1$    & 0.83 & \textbf{0.94} & 0.89 & \textbf{0.94} & \textbf{0.94} \\
Contextual Relevancy $\rightarrow 1$ & \textbf{0.72} & 0.58 & 0.55 & 0.51 & 0.47 \\
Faithfulness $\rightarrow 1$         & 0.96 & \textbf{1.00} & 0.97 & \textbf{1.00} & 0.97 \\
Answer Relevancy $\rightarrow 1$     & 0.87 & \textbf{0.94} & 0.93 & \textbf{0.94} & \textbf{0.94} \\
\bottomrule
\end{tabular*}

\end{table}

\subsection{Qualitative Retrieval Analysis}
\label{sec:res-qualitative}

\begingroup
\linespread{1.10}\selectfont
\setlength{\parskip}{0pt}

The following cases compare the clinical information need with the retrieved evidence. Excerpts are taken from the ranked passages, rather than from the reference answers.
\par\vspace{0.85\baselineskip}

\noindent\textbf{Q01 --- Flat epithelial atypia (successful; rank 1).}\par
\vspace{0.30\baselineskip}
\noindent\textit{Query:}\par
\vspace{0.12\baselineskip}
{\leftskip=1.5em\noindent
``Is diagnostic excision generally indicated for flat epithelial atypia alone when radiologic-pathologic concordance is confirmed?''\par}
\vspace{0.35\baselineskip}
\noindent\textit{Retrieved:}\par
\vspace{0.12\baselineskip}
{\leftskip=1.5em\noindent
``Provided radiologic-pathologic concordance is confirmed, diagnostic excision is generally not indicated after a diagnosis of FEA'' (Nakhlis et al., 2026).\par}
\vspace{0.40\baselineskip}
\noindent The passage directly states the management recommendation and preserves the concordance requirement.\par
\vspace{0.95\baselineskip}

\noindent\textbf{Q02 --- Radial scar/complex sclerosing lesion (successful; rank 1).}\par
\vspace{0.30\baselineskip}
\noindent\textit{Query:}\par
\vspace{0.12\baselineskip}
{\leftskip=1.5em\noindent
``Can surgical excision generally be avoided for an adequately sampled, concordant radial scar or complex sclerosing lesion without atypia?''\par}
\vspace{0.35\baselineskip}
\noindent\textit{Retrieved:}\par
\vspace{0.12\baselineskip}
{\leftskip=1.5em\noindent
``Routine excision is not required if pathology is benign concordant with radiology assessment'' (Eisen et al., 2026; section on pure radial scars/complex sclerosing lesions without atypia).\par}
\vspace{0.40\baselineskip}
\noindent The retrieved passage supports an affirmative answer to the Yes/No question, indicating that routine surgical excision is not required for radiologic--pathologic concordant lesions without atypia. However, sampling adequacy is not explicitly addressed in the quoted excerpt. Thus, although the passage was accepted as a successful retrieval in the benchmark, the excerpt alone does not establish explicit coverage of every clinical qualifier in the question.\par
\vspace{0.95\baselineskip}

\noindent\textbf{Q03 --- Atypical lobular hyperplasia (successful; rank 1).}\par
\vspace{0.30\baselineskip}
\noindent\textit{Query:}\par
\vspace{0.12\baselineskip}
{\leftskip=1.5em\noindent
``Is excision generally required for atypical lobular hyperplasia when imaging is concordant and no other lesion is present?''\par}
\vspace{0.35\baselineskip}
\noindent\textit{Retrieved:}\par
\vspace{0.12\baselineskip}
{\leftskip=1.5em\noindent
``Generally, ALH does not require excision if breast imaging has ruled out any other lesions and is reported as concordant'' (Eisen et al., 2026).\par}
\vspace{0.40\baselineskip}
\noindent The passage directly addresses the management decision and both clinical qualifiers.\par
\vspace{0.95\baselineskip}

\noindent\textbf{Q09 --- Pseudoangiomatous stromal hyperplasia (Top-3 failure).}\par
\vspace{0.30\baselineskip}
\noindent\textit{Query:}\par
\vspace{0.12\baselineskip}
{\leftskip=1.5em\noindent
``Which biopsy technique is preferred for diagnosing pseudoangiomatous stromal hyperplasia?''\par}
\vspace{0.35\baselineskip}
\noindent\textit{Retrieved (rank 1):}\par
\vspace{0.12\baselineskip}
{\leftskip=1.5em\noindent
``Core needle biopsy is indicated for diagnosis and must be concordant with imaging and clinical findings'' (Estes et al., 2020).\par}
\vspace{0.35\baselineskip}
\noindent\textit{Accepted evidence (rank 7):}\par
\vspace{0.12\baselineskip}
{\leftskip=1.5em\noindent
``Core needle biopsy is preferred over fine needle aspiration (FNA) or excisional biopsy.''\par}
\vspace{0.40\baselineskip}
\noindent Although relevant to PASH diagnosis, the rank-1 passage does not compare biopsy techniques. The required comparative evidence was retrieved at rank 7, outside the Top-3 cutoff.\par

\endgroup

\section{Discussion}
\label{sec:discussion}

Retrieval performance depended on both the selected configuration and the coverage of the annotated ground truth. Increasing generation-context depth also produced different effects across evaluation metrics, without establishing a consistently superior setting. These findings characterize a small development benchmark and do not constitute independent clinical validation.

\subsection{Technical Development}
\label{sec:dis-technical}

The selected configuration (384-token chunks, 20\% overlap, Hybrid RRF with cross-encoder reranking) retrieved accepted evidence within the top three results for 17 of 18 queries (Recall@3 = 0.94). In this benchmark, accepted evidence chunks were evaluated as alternative individual targets, retaining the maximum score per query and metric. Recall@3 therefore indicates whether at least one accepted alternative appears in the top three, rather than the proportion of all relevant chunks retrieved. This approach reduces dependence on a single annotated chunk but does not establish invariance to overlapping evidence.

Ground-truth curation increased the number of evidence spans from 25 to 37 and accepted chunk targets from 36 to 51. Because retrieval rankings remained unchanged, changes in evaluation scores reflect expanded annotation coverage rather than improved retrieval. However, the audit examined retrievals from the selected configuration, potentially favoring that system in subsequent comparisons.

Hybrid RRF increased R-Precision to 0.78, compared with 0.61 for BM25 and 0.56 for semantic retrieval, while Recall@3 remained at 0.89. Adding the cross-encoder reranker increased Recall@3, R-Precision, MRR@3, and nDCG@3 to 0.94. Segmentation also influenced performance: 384-token chunks achieved the highest MRR@3 (0.89) and nDCG@3 (0.90) in the size sweep, while 20\% overlap produced the highest R-Precision, MRR@3, and nDCG@3 among the evaluated overlap settings. The lower performance at 15\% overlap indicates that these effects were not monotonic.

Generation performance varied with context depth. Contextual Relevancy decreased from 0.72 at \(K=1\) to 0.47 at \(K=5\), whereas Contextual Precision was highest at \(K=5\) (0.91). Contextual Recall reached 0.94 at \(K=2\), \(K=4\), and \(K=5\), while Faithfulness remained between 0.96 and 1.00. Although \(K=5\) achieved the highest BLEU-1 through BLEU-4 and ROUGE-1/ROUGE-2 scores, ROUGE-L was highest at \(K=1\). Increasing context depth therefore did not consistently improve all measures of answer quality.

The qualitative cases illustrate the distinction between benchmark success and evidence sufficiency. In Q02, the retrieved passage supported an affirmative answer but did not explicitly address sampling adequacy in the quoted excerpt. In Q09, the rank-one passage addressed the relevant biopsy technique without establishing the requested comparison; the accepted comparative evidence appeared only at rank seven. These cases demonstrate that topical relevance and agreement with annotated targets do not necessarily establish coverage of every clinical qualifier.

Several limitations constrain interpretation. The corpus contained 18 documents and 18 simple, direct queries, with the same benchmark used for configuration selection and evaluation. A single query changes Recall@3 by approximately 0.06, making differences between configurations sensitive to individual cases. Generation quality was evaluated using an LLM judge and lexical metrics without human validation. Additionally, noisy chunks were excluded from ground-truth candidates, and the corpus did not include the discordant guidance recommending vacuum-assisted excision as the default for B3 lesions. Generalization to independent questions, more complex evidence requirements, and conflicting clinical sources remains untested.

\subsection{Clinical Implications}
\label{sec:dis-implications}

Following image-guided biopsy, breast radiologists assess radiologic--pathologic concordance and determine whether surgical excision, imaging surveillance, or return to screening is appropriate. These decisions depend on lesion-specific recommendations and clinical qualifiers, including atypia and sampling adequacy. Retrieving supporting passages with identifiable sources could facilitate verification against the relevant literature, although this potential benefit was not evaluated clinically.

For clinical decision support, retrieving an accepted evidence target is insufficient if the passage omits conditions necessary for the management recommendation. The evidence-aware evaluation identifies recognized supporting passages but does not independently establish their clinical sufficiency. Expert assessment is therefore required to determine whether retrieved evidence preserves the relevant conditions and supports the intended interpretation.

A planned reader study will compare a breast imaging radiologist's assessment of retrieved passages with the automated ranking. Disagreements may inform future ground-truth revisions and evaluation of clinical evidence sufficiency. The system is intended to support evidence retrieval rather than replace clinical judgment, and its effectiveness in clinical workflows remains to be established.

\section{Conclusion and Future Work}
\label{sec:conclusion-future-work}

The RAG-MAMA evaluation illustrates how retrieval performance depends on both system configuration and the coverage of relevance annotations. On the 18-query development benchmark, hybrid RRF with cross-encoder reranking retrieved at least one accepted evidence alternative within the top three results for 17 queries (Recall@3 = 0.94). Ground-truth curation expanded canonical evidence spans from 25 to 37 and accepted chunk targets from 36 to 51 without changing retrieval rankings. Generation performance also varied across context depths, with no single setting consistently maximizing all metrics. These findings are limited to a small development benchmark and do not establish clinical evidence sufficiency or clinical effectiveness.

The immediate priorities are a blinded audit of candidate passages pooled across multiple retrievers and evaluation on independent questions without further configuration tuning. A planned reader study will ask a fellowship-trained breast imaging radiologist to assess and rank the top five retrieved chunks for up to 20 stratified queries, flag irrelevant passages, and evaluate whether the evidence preserves the clinical qualifiers needed for management decisions. Human assessment of generated-answer faithfulness and correctness is also needed. The benchmark should subsequently be extended to comparative questions and questions requiring complementary evidence from multiple passages.

Further technical work will compare fixed-size chunking with semantic and section-aware segmentation, particularly for evidence spanning paragraphs, tables, and quantitative results. We also plan to investigate down-weighting noisy chunks instead of excluding them, including fixed versus proportional penalties, document-position effects, and query-dependent rules. Graph-based retrieval will be explored for relational and multi-hop evidence needs.


\clearpage
\appendix
\section{Supplementary Material}
\label{sec:supplementary}

This supplementary material reports the exploratory query-formulation analysis, including its experimental setup and descriptive retrieval results.

\subsection{Query Formulation Sensitivity Analysis}
\label{sec:supp-query-formulation}

\subsubsection{Experimental Setup}

Query-formulation sensitivity was evaluated across seven conditions (Short, Medium, Long, Canonical, Paraphrased, Yes/No, and WH), with 18 queries per condition. Fixed variants were authored through Codex (recorded session model: \texttt{gpt-5.6-luna}; \texttt{gpt-6.1-sol} for eight additional indirect forms). Medium and Canonical retained the original queries; mean Short, Medium, and Long lengths were 11, 13, and 18 words, respectively. Paraphrases varied wording or syntax while aiming to preserve the original answers and evidence. Eight Yes/No variants were indirect requests containing WH clauses, limiting a strict binary-versus-open question comparison.

All 126 variants were evaluated on the selected 384-token index (77-token overlap; 757 chunks) using the curated ground truth and the final ablation's retrieval settings and scoring protocol. GPT-2 perplexity was reused from prior evaluations of identical query texts and averaged by condition, without new perplexity inference. Answer generation and LLM-based judging were disabled. Analyses were descriptive.

\subsubsection{Results}

Table~\ref{tab:supp-query-formulation} summarizes retrieval performance across seven query-formulation conditions. Medium-length queries achieved the highest R-Precision (0.89), MRR@3 (0.92), and nDCG@3 (0.91) among the length variants. Canonical and paraphrased queries performed similarly, with R-Precision values of 0.89 and 0.88, respectively. Yes/No questions outperformed WH questions in Recall@3 (0.89 versus 0.83), R-Precision (0.87 versus 0.84), and nDCG@3 (0.91 versus 0.85), while WH questions achieved slightly higher MRR@3 (0.92 versus 0.91). Neither query length nor perplexity showed a consistent monotonic association with retrieval performance. These differences are descriptive and were not tested for statistical significance.

\renewcommand{\thetable}{S\arabic{table}}
\renewcommand{\theHtable}{supp.\arabic{table}}
\setcounter{table}{0}

\begin{table}[htbp]
\centering
\small
\setlength{\tabcolsep}{3pt}
\caption{Query formulation sensitivity analysis on the 18-query benchmark. $\rightarrow 1$ indicates the ideal retrieval score; $\downarrow$ indicates lower perplexity. Bold values identify the best retrieval scores within each formulation category before rounding.}
\label{tab:supp-query-formulation}

\begin{tabular*}{\textwidth}{@{\extracolsep{\fill}}lccccc@{}}
\toprule
Condition & Mean PPL $\downarrow$ & Recall@3 $\rightarrow 1$ & R-Precision $\rightarrow 1$ & MRR@3 $\rightarrow 1$ & nDCG@3 $\rightarrow 1$ \\
\midrule
Length: Short        & 300.05 & 0.85 & 0.80 & 0.86 & 0.84 \\
Length: Medium       & 231.77 & \textbf{0.89} & \textbf{0.89} & \textbf{0.92} & \textbf{0.91} \\
Length: Long         & 134.84 & 0.86 & 0.75 & 0.88 & 0.85 \\
\midrule
Wording: Canonical   & 231.77 & \textbf{0.89} & \textbf{0.89} & \textbf{0.92} & \textbf{0.91} \\
Wording: Paraphrased & 302.03 & 0.88 & 0.88 & \textbf{0.92} & 0.90 \\
\midrule
Form: Yes/No         & 234.39 & \textbf{0.89} & \textbf{0.87} & 0.91 & \textbf{0.91} \\
Form: WH             & 169.93 & 0.83 & 0.84 & \textbf{0.92} & 0.85 \\
\bottomrule
\end{tabular*}

\end{table}

\end{document}